\documentclass[runningheads]{llncs}

\usepackage[T1]{fontenc}
\usepackage{graphicx}
\usepackage{booktabs}
\usepackage{amsmath,amssymb}
\usepackage{adjustbox}
\usepackage{float}
\usepackage{algorithm}
\usepackage{algpseudocode}
\usepackage{multirow}
\usepackage{tabularx}
\usepackage{array}
\usepackage{caption}
\usepackage{color}
\usepackage{url}
\usepackage[hidelinks]{hyperref}

\begin{document}
\title{SG-Layout: Structured Scene Graph--Guided Layout Generation with LLMs}
\titlerunning{SG-Layout}

\author{
Junsheng Wang \and
Chao Chen\thanks{Corresponding author.} \and
Mengying Xie \and
Mingyan Li \and
Fuqiang Gu
}

\authorrunning{J. Wang et al.}

\institute{
Chongqing University, Chongqing, China\\
\email{\{csjunshengwang,cschaochen,xiemy,limy2021,gufq\}@cqu.edu.cn}
}

\maketitle

\begin{abstract}
Understanding and generating spatially coherent layouts from natural language remains a fundamental yet challenging task for large language models (LLMs).
Existing LLMs often struggle to capture explicit geometric relationships and structural dependencies between objects.
To address this issue, we propose \textbf{SG-Layout}, a graph-guided layout generation framework that explicitly incorporates structured spatial knowledge into LLMs.
SG-Layout follows a two-stage training paradigm: (1) a graph–language feature alignment stage, where a relational graph encoder and a projector are trained to map scene-graph embeddings into the LLM’s linguistic space; and (2) an instruction tuning stage, where LoRA-based adapters enable efficient fine-tuning for instruction-driven layout generation while keeping the backbone frozen.
We evaluate SG-Layout on image layout generation, indoor scene synthesis and robotic object rearrangement tasks.
Experimental results show that SG-Layout improves spatial reasoning accuracy and geometric consistency over the compact open-source backbone, with particularly clear advantages in relation-dense and compositionally complex scenes.
These results highlight the effectiveness of graph-structured feature alignment for enhancing controllable layout generation.
\end{abstract}

\section{Introduction}
\label{sec:intro}

Layout generation refers to the process of producing structured representations of object arrangements—such as positions, sizes, and orientations—based on input conditions including textual descriptions, semantic labels, or task-specific instructions \cite{jyothi2019layoutvae,li2019layoutgan}. Serving as a bridge between high-level task planning and low-level visual generation or robotic execution, layout generation has become an essential intermediate step across multiple fields. In computer vision, it is often employed as a precursor to text-to-image synthesis, where precise spatial arrangements enhance semantic consistency in generated images \cite{zhao2020layout2image,gupta2021layouttransformer,li2021image}. Within embodied intelligence, layout generation provides actionable spatial configurations for robotic manipulation and object rearrangement tasks \cite{paschalidou2021atiss,li2019grains,ding2023task,vidanapathirana2021plan2scene,wang2021sceneformer}.

\begin{figure}[!t]
\centering
\includegraphics[width=\linewidth]{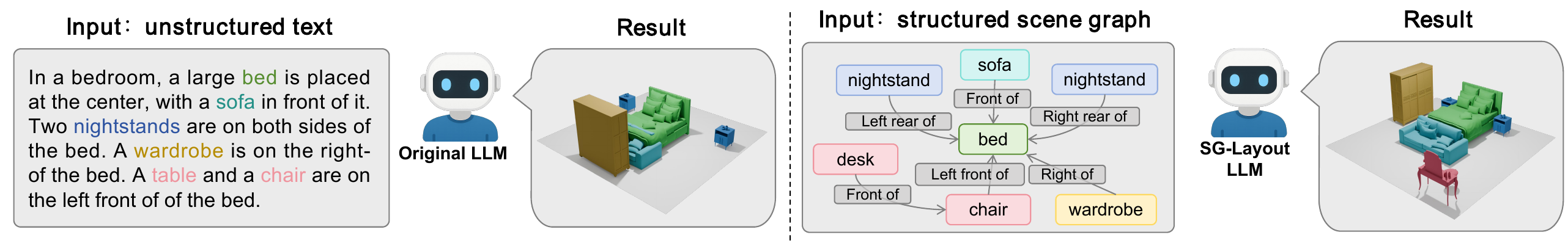}
\caption{Scene graph guides LLM for layout generation.}
\vspace{-20pt}
\label{fig_1}
\end{figure}

Recent advances in large language models (LLMs) have demonstrated great potential for layout generation, as their strong semantic understanding and reasoning abilities enable them to translate natural language instructions into structured spatial configurations \cite{phung2024grounded,wu2024self,gani2023llm,ran2025direct,sun2025layoutvlm,qu2023layoutllm}. To further mitigate the limitations of data scarcity and insufficient spatial reasoning, retrieval-augmented generation (RAG) techniques have been introduced. Representative frameworks such as LayoutGPT \cite{feng2023layoutgpt} and SKE-Layout \cite{wang2025ske} incorporate external layout knowledge bases, retrieving task-relevant examples as contextual guidance for LLM-driven layout generation. Despite these efforts, current LLM-based approaches still face a fundamental limitation: natural language alone is inherently ambiguous and lacks the structural rigor required to express spatial layouts. Moreover, as scenes grow more complex, models struggle to ground textual cues into metrically correct object sizes, positions, and poses.
These limitations highlight the need to move beyond purely text-based representations. Scene graphs provide explicit, compositional modeling of objects and their spatial relations, while LLMs contribute strong semantic understanding, instruction following, and open-world generalization. Integrating scene graphs into the LLM-driven layout generation pipeline converts ambiguous language into structured guidance, tightening alignment between user instructions and produced layouts and yielding better layouts in both 2D and 3D scenes, as shown in Fig.~\ref{fig_1}.

Achieving high-precision graph-guided layout generation requires addressing two fundamental challenges \cite{perozzi2024let,fatemi2024talk}.
First, LLMs are text-native models and are not well suited to directly consume structured inputs such as scene graphs. Bridging this modality gap requires a mechanism to encode scene graphs into embeddings that are semantically aligned with the LLM's latent space, enabling the model to interpret and reason over structured spatial knowledge in a language-compatible form.
Second, even with semantically aligned representations, effectively injecting graph-based information into the layout generation process remains non-trivial.  It demands training strategies that support efficient adaptation and structural generalization, without disrupting the pretrained language backbone or incurring prohibitive computational costs.

To address these challenges, we propose the \textbf{\underline{S}cene \underline{G}raphs-\underline{L}ayouts (SG-Layout)} framework,
a two-stage graph-guided approach for text-conditioned layout generation.
In the first stage, \emph{graph–language feature alignment}, natural language instructions are parsed into structured scene graphs, where nodes represent objects and their attributes, and edges encode semantic and spatial relations.
A relational graph encoder and a projector are trained to align graph embeddings with the LLM’s linguistic latent space, allowing the model to interpret structured spatial information in a unified representation.
In the second stage, \emph{instruction tuning}, the graph encoder and projector are frozen, and LoRA-based \cite{hu2021loralowrankadaptationlarge} adapters are introduced to efficiently fine-tune the LLM for instruction-driven layout generation.
This non-invasive injection keeps the backbone stable and updates only a small fraction of parameters, yielding parameter-efficient adaptation without full-parameter fine-tuning.
Together, the two stages produce more accurate and semantically consistent layouts under complex spatial constraints while preserving the pretrained backbone parameters.

The main contributions are summarized as follows:
\begin{itemize}
    \item We introduce \textbf{SG-Layout}, a graph-guided layout generation framework that bridges structured spatial representations and language modeling through a two-stage training paradigm.
    \item We design a graph–language alignment mechanism that maps relational scene-graph embeddings into the LLM’s token space, and a LoRA-based instruction tuning strategy that efficiently adapts the model for layout generation without full fine-tuning.
    \item We conduct evaluations across \emph{image layout generation}, \emph{indoor scene synthesis}, and \emph{object rearrangement}, showing that graph conditioning provides the largest gains in relation-dense and compositionally complex scenes, while improving overall spatial consistency over the same Qwen3-8B backbone.
\end{itemize}

\section{Related Work}
\label{sec:relatedwork}

\subsection{Image Layout Generation}
Image layout generation aims to produce structured object arrangements for downstream visual synthesis. Early deep learning methods, such as LayoutGAN~\cite{li2019layoutgan}, LayoutVAE~\cite{jyothi2019layoutvae}, and LayoutTransformer~\cite{gupta2021layouttransformer}, are usually trained under fixed categories or partially specified layout conditions, which limits their flexibility for free-form instructions. Recent LLM-based approaches attempt to bridge this gap by mapping natural language prompts into structured layouts. LayoutGPT~\cite{feng2023layoutgpt} uses an LLM planner to generate CSS-like spatial descriptors and hierarchical layout plans, while subsequent methods improve object grounding through attention refocusing~\cite{phung2024grounded}, LLM-in-the-loop correction~\cite{wu2024self}, compositional blueprints~\cite{gani2023llm}, or lightweight semantic adapters~\cite{zhong2023adapter}. Automated correction strategies have also been explored to mitigate spatial errors during or after generation~\cite{pan2023automatically}. SKE-Layout~\cite{wang2025ske} further introduces retrieval-augmented spatial knowledge to guide LLM-based planning. However, these methods still mainly rely on text prompts or retrieved textual examples, and thus may struggle with complex object-to-object relations. In contrast, SG-Layout explicitly injects structured scene-graph embeddings into the LLM, enabling more direct modeling of compositional spatial constraints.

\subsection{Indoor Scene Synthesis}
Indoor scene synthesis focuses on generating plausible 3D room layouts that satisfy both semantic and geometric constraints. Existing methods can be broadly divided into unconditional and conditional generation. Unconditional methods learn scene distributions from large-scale datasets and synthesize complete rooms from object sets, such as ATISS~\cite{paschalidou2021atiss} and GRAINS~\cite{li2019grains}. Although these methods can generate diverse and plausible scenes, they are not designed to faithfully follow user-specified instructions. Conditional methods incorporate additional constraints, including floor plans, partial layouts, semantic descriptions, or scene graphs. Representative examples include SceneFormer~\cite{wang2021sceneformer}, DiffuScene~\cite{tang2024diffuscene}, Graph-to-3D~\cite{dhamo2021graph}, LayoutVLM~\cite{sun2025layoutvlm}, DirectLayout~\cite{ran2025direct}, and LayoutGPT~\cite{feng2023layoutgpt}. These methods improve controllability from different perspectives, but complex spatial reasoning and metrically grounded object placement remain challenging, especially when instructions involve multiple interacting objects and relations. Recent graph-conditioned methods, such as InstructScene~\cite{lin2024instructscene} and EchoScene~\cite{zhai2024echoscene}, also demonstrate the value of explicit relational structure for 3D scene generation. SG-Layout follows this direction but formulates scene graphs as graph tokens aligned with an LLM, allowing structured spatial priors to guide layout generation in a parameter-efficient manner.

\subsection{Large Language Models for Embodied Spatial Reasoning}
LLMs have shown emerging potential for embodied spatial reasoning and task planning. For example, Language Models as Zero-Shot Planners~\cite{huang2022language} demonstrates that LLMs can decompose high-level instructions into executable plans, while LLM-GROP~\cite{ding2023task} extracts commonsense object-placement knowledge from LLMs for manipulation and layout reasoning. However, such methods mainly depend on implicit language representations, which may be insufficient for accurately modeling explicit spatial dependencies. Graph-structured representations provide a more compositional interface for objects and their relations. SayPlan~\cite{rana2023sayplan} maps high-level instructions to scene graphs for scalable robotic planning, and SG-Bot~\cite{zhai2024sg} uses a coarse-to-fine scene-graph pipeline for object rearrangement. These works suggest that explicit structure can improve spatial organization, while SG-Layout further integrates such structure into an LLM-based layout generation framework through graph-language alignment and LoRA-based adaptation.

\subsection{Graph-Structured Representations for LLMs}
Recent studies have explored graph-structured representations as controllable inputs for language models. Let Your Graph Do the Talking~\cite{perozzi2024let} serializes graph data into token sequences so that LLMs can directly process structured information, and GraphGPT~\cite{tang2024graphgpt} aligns LLMs with graph inputs for graph-conditioned instruction following and reasoning. These works mainly focus on graph understanding or language-centric reasoning tasks, where outputs are typically answers, explanations, or decisions. SG-Layout differs in both objective and evaluation: it uses scene graphs as an explicit spatial interface for layout synthesis, conditions generation on graph embeddings, and outputs metrically grounded geometric attributes such as object boxes, positions, and poses. The resulting layouts are evaluated not only by linguistic correctness but also by spatial consistency and physical feasibility, including out-of-bound and collision metrics.
\section{Method}
\label{sec:method}

\begin{figure}[!h]
\centering
\includegraphics[width=\linewidth]{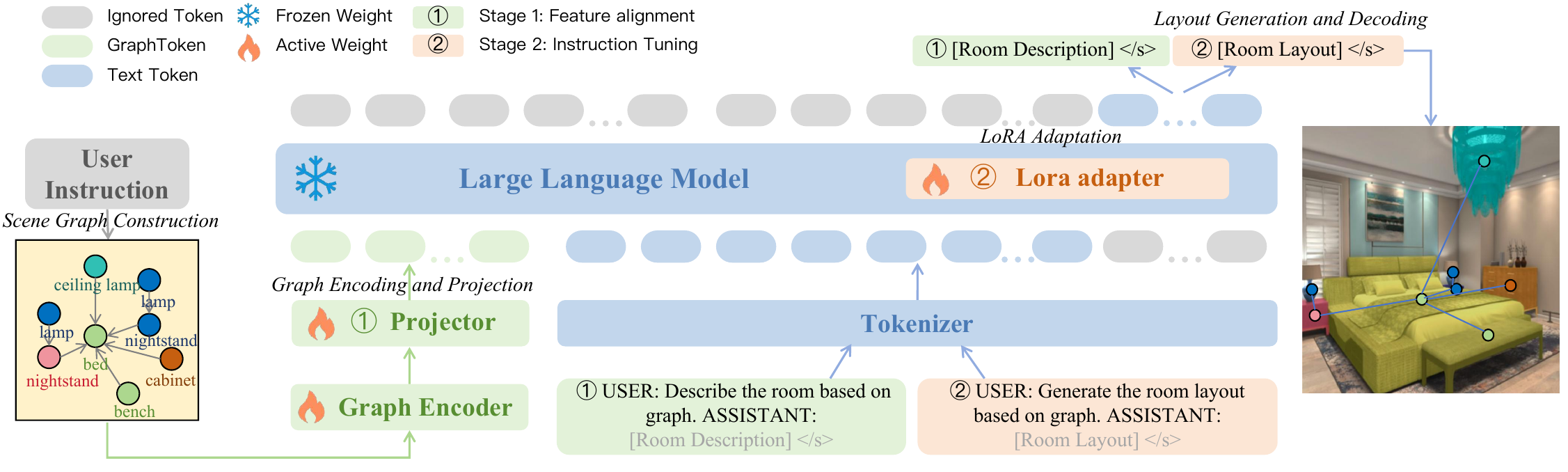}
\caption{Framework of SG-Layout. User instructions are parsed into scene graphs, encoded into graph embeddings, and injected into an LLM via a graph encoder and LoRA adapter to guide layout generation.}
\vspace{-20pt}
\label{fig_2}
\end{figure}

As illustrated in  Fig.~\ref{fig_2}, \textbf{SG-Layout} proposes a scene graph–guided LLM framework for layout generation. First, user instructions in natural language are parsed and combined with scene graphs generated by GPT-4o under rule guidance, in order to capture objects and their spatial relations. The scene graph is then encoded by a \emph{Graph Encoder}, and the resulting structured representations are projected by a \emph{Projector} into Graph Tokens that are compatible with the input space of the language model. These graph tokens are concatenated with text tokens obtained by tokenizing the user instructions, forming a unified input sequence to the pretrained LLM. To efficiently adapt the model to task-specific requirements, we insert \emph{LoRA adapters} into the LLM, freezing the original weights while training only a small number of parameters. Finally, by integrating semantic information with structural constraints, the model outputs a textual description of the room layout, which is further decoded into 2D or 3D layouts. Overall, the framework effectively combines explicit spatial constraints from scene graphs with the semantic generalization capability of LLMs, enabling efficient and plausible layout generation across 2D and 3D scenarios.

\subsection{Problem Settings}

We study the task of layout generation from natural language instructions, where the goal is to produce a structured layout that satisfies both semantic and spatial constraints. Formally, we define the problem as follows.
\begin{definition}
\textbf{User instruction.}
A user instruction is a natural-language text denoted by $u$, which specifies concrete objects and their pairwise spatial relations. It serves as the textual condition for constructing the scene graph $\hat{G}$ and generating the layout $L$.
\end{definition}
\begin{definition}
\textbf{Scene graph representation.}
A scene graph is denoted as:
\begin{equation}
    \hat{G} = (V, E),
\end{equation}
where $V = \{v_1, v_2, \ldots, v_n\}$ represents the set of object nodes. Each node $v_i = (id_i, p_i)$ encodes the object identity $id_i$ and its attribute information $p_i$. The edge set $E = \{e_{ij}\}$ describes the spatial relations between objects, where $e_{ij} = r_{ij}$ and $e_{ij} \neq e_{ji}$. Each relation $r_{ij}$ belongs to a predefined set of spatial relations:
\begin{equation}
    \mathcal{R} = \{\text{left\_of}, \text{right\_of}, \text{front\_of}, \text{behind}, \text{above}, \text{below}\}.
\end{equation}
In practice, $r_{ij} \in \{0,1\}^{|\mathcal{R}|}$ is represented as a multi-hot vector indicating the relation types contained in the edge.
\end{definition}
\begin{definition}
\textbf{Layout representation.}
A layout is represented as:
\begin{equation}
    L = \{o_1, o_2, \ldots, o_n\},
\end{equation}
where each object is defined as:
\begin{equation}
    o_i = \{id_i, x_i, y_i, z_i, w_i, h_i, d_i\}.
\end{equation}
Here $id_i$ denotes the object category, $(x_i, y_i, z_i)$ its 3D position, and $(w_i, h_i, d_i)$ its width, height, and depth.
\end{definition}
\textbf{Problem Statement}. Given a natural language instruction $u$, the objective is to generate a layout $L$ such that the spatial arrangement of objects satisfies the semantic description and adheres to the structural constraints encoded in the scene graph $\hat{G}$.

\subsection{Dataset Construction}
During training, we construct \emph{Instruction--Scene Graph--Layout} triplets $(u,\hat{G},L)$ to provide supervised signals for graph-augmented generation. For 2D layout generation, we build an \emph{MSCOCO-Subset} by filtering MSCOCO instances according to category and bounding-box size, removing tiny or task-irrelevant objects, and grouping the remaining layouts by object count. A VLM is used to extract candidate scene graphs and suitability decisions, followed by manual verification to correct object lists and pairwise relations when necessary. For 3D indoor scene synthesis, we use bedroom and living-room scenes from 3D-FRONT/3D-FUTURE. We extract furniture categories, 3D positions, sizes, rotations, and room dimensions from the original annotations, and query a vision-language model with the layout summary and rendered room image to obtain an oriented scene graph. For object rearrangement, we use the relational subset of SK-Dataset containing qualitative spatial instructions. The resulting dataset is represented as
\begin{equation}
    \mathcal{D} = \{(u_i, \hat{G}_i, L_i)\}_{i=1}^N.
\end{equation}

\subsection{Layout Generation Pipeline}
\paragraph[Scene Graph Construction.]{\textbullet\enspace Scene Graph Construction.}
We parse user instructions into structured scene graphs with the help of GPT-4o. Given a natural language input $u \in U$, the model generates the graph representation $\hat{G} = (V, E)$. This process allows ambiguous natural language to be transformed into a symbolic form that captures objects and their spatial relations.
\paragraph[Graph Encoding and Projection.]{\textbullet\enspace Graph Encoding and Projection.}
Once the scene graph $\hat{G}$ is constructed, we encode it to obtain embeddings that reflect both semantic and structural dependencies. We adopt a relational graph transformer (RGT) \cite{dwivedi2025relational} as the default encoder:
\begin{equation}
    Z_G = f_{\text{RGT}}(\hat{G}),
\end{equation}
where $Z_G$ contains node- and relation-aware representations. Since these embeddings live in a graph space that is not directly compatible with the token space of the LLM, we introduce a projector:
\begin{equation}
    T_G = f_{\text{proj}}(Z_G),
\end{equation}
which maps the graph embeddings into the same latent space as user instruction text tokens. This alignment ensures that spatial knowledge extracted from graphs can be fused with text representations in subsequent stages.
\vspace{-5pt}
\paragraph[Large Language Model with LoRA Adaptation.]{\textbullet\enspace Large Language Model with LoRA Adaptation.}
We encode the user instruction into text tokens via a tokenizer:
\begin{equation}
    T_u = f_{\text{Tokenizer}}(u).
\end{equation}
The text tokens $T_u$ and the projected graph tokens $T_G$ are concatenated to form a unified sequence:
\begin{equation}
    T = [T_G; T_u].
\end{equation}
This fused representation is fed into a large language model enhanced with LoRA adapters:
\begin{equation}
    H = f_{\text{LLM}}(T;\theta,\Delta\theta_{\text{LoRA}}),
\end{equation}
where \(\theta\) denotes the frozen backbone parameters and \(\Delta\theta_{\text{LoRA}}\) denotes the trainable low-rank updates.

In our implementation, we adopt the Qwen3-8B model \cite{yang2025qwen3} as the backbone LLM. LoRA modules allow the model to be efficiently adapted to layout generation tasks by training only a small number of low-rank parameters while keeping the backbone frozen. This design achieves a balance between efficiency and flexibility, enabling the LLM to incorporate spatial knowledge without sacrificing scalability.
\vspace{-5pt}
\paragraph[Layout Generation and Decoding.]{\textbullet\enspace Layout Generation and Decoding.}
Finally, the LLM autoregressively predicts a serialized layout sequence. Each object is represented by its category and geometric attributes, and the predicted sequence is parsed back into a structured layout:
\begin{equation}
    \hat{L} = f_{\text{Decode}}(H) = \{o_i = (id_i, x_i, y_i, z_i, w_i, h_i, d_i)\}_{i=1}^n .
\end{equation}
For 2D layout generation, the same schema is used after omitting the unused depth dimension. This text-to-structure decoding keeps the output compatible with the LLM vocabulary while preserving the object-level coordinates, sizes, and orientations required for geometric evaluation.

\subsection{Loss Function}
We supervise layout generation using token-level cross-entropy on the serialized ground-truth layout sequence with teacher forcing, conditioning on both the user instruction and scene graph:
\begin{equation}
    \mathcal{L}_{\text{gen}} = -\sum_{t=1}^{|L|}\log P(l_t \mid l_{<t}, u, \hat{G}).
\end{equation}
This objective encourages the model to generate layout tokens that are faithful to the input instruction, the graph-encoded pairwise relations, and the ground-truth geometric attributes.

\subsection{Two Stage Training Strategy}

We adopt a two-stage training paradigm while keeping the LLM backbone frozen; only the graph-side modules or the LoRA adapters are optimized at each stage.

At the first stage, we align scene-graph representations with the language embedding space while keeping both the LLM and LoRA adapters frozen. Given a scene graph $\hat{G}$, we encode it with a relational graph encoder and project it into the LLM token space to obtain graph tokens $T_G$. A brief prompt (e.g., ``Describe the room.'') is tokenized into $T_u$, and the concatenated sequence $[T_G; T_u]$ is fed to the frozen LLM. The model is supervised to produce a concise scene description; minimizing token-level cross-entropy on this output aligns graph embeddings with the language space without modifying the LLM weights.

At the second stage, we perform instruction tuning for layout generation with the LoRA adapters made trainable, while the graph encoder and projector are kept frozen. Given the user instruction $u$ and the precomputed graph tokens $T_G$, we form $[T_G; T_u]$ and autoregressively predict the target layout sequence $L$. A standard generation loss (token-level cross-entropy against the ground-truth layout sequence) updates only the LoRA parameters, yielding parameter-efficient adaptation while preserving the pretrained backbone.

\section{Experiment}
\label{sec:experiment}

In this section, we present extensive empirical results of the proposed \textbf{SG-Layout} model and several competitive baseline methods.
Across representative tasks involving spatial reasoning, including \emph{image layout generation, indoor scene synthesis and robotic object rearrangement}, our proposed SG-Layout generally improves spatial understanding and reasoning ability, with the clearest advantages in complex and relation-dense settings. Compared to the original LLM backbone, SG-Layout achieves notably higher task success rates, confirming the effectiveness of incorporating graph-structured spatial knowledge.

\subsection{Experimental Setup}
\paragraph[Implementation Details.]{\textbullet\enspace Implementation Details.}
We adopt Qwen3-8B as the backbone LLM, which provides strong instruction-following and text generation capabilities.
The graph encoder is implemented using RGT, which extracts semantic and geometric features from scene graphs and encodes their spatial relationships (e.g., \emph{left of}, \emph{in front of}).

\begin{table}[t]
\centering
\vspace{-6pt}
\caption{Baselines and comparison protocols. Controlled ablations use the same Qwen3-8B backbone and task data; reference baselines follow their standard settings.}
\adjustbox{max width=\linewidth}{
\begin{tabular}{lll}
\toprule
\textbf{Category} & \textbf{Task} & \textbf{Models} \\
\midrule
Controlled ablations & All tasks & Qwen3; Qwen3+LoRA; SG-Layout \\
\midrule
Reference baselines & 2D image layout & LayoutGPT (GPT-3.5/GPT-4) \cite{feng2023layoutgpt}; SKE-Layout \cite{wang2025ske} \\
Reference baselines & 3D indoor scene & DiffuScene \cite{tang2024diffuscene}; InstructScene \cite{lin2024instructscene}; GPT-4; SKE-Layout \\
Reference baselines & Object rearrangement & LLM-GROP \cite{ding2023task}; SKE-Layout \\
\bottomrule
\end{tabular}
}
\label{tab:baselines}
\vspace{-12pt}
\end{table}

\begin{table}[t]
\centering
\caption{Overview of datasets and data scales used in our experiments.}
\setlength{\tabcolsep}{5pt}
\begin{tabularx}{\linewidth}{l l X}
\toprule
\textbf{Dataset} & \textbf{Task category} & \textbf{Data scale} \\
\midrule
NSR-1K & Numerical + spatial reasoning & 39,436 \\
MSCOCO-Subset & 2 / 4 / 6 / 8 objects & $4\times1000$ \\
3D-FRONT / 3D-FUTURE & Bedroom scenes & 4,041 \\
3D-FRONT / 3D-FUTURE & Living-room scenes & 813 \\
SK-Dataset & 2 / 4 / 6 / 8 objects & $4\times1000$ \\
\bottomrule
\end{tabularx}
\label{tab:data_scale}
\vspace{-8pt}
\end{table}

\paragraph[Baselines.]{\textbullet\enspace Baselines.}
Table~\ref{tab:baselines} separates controlled ablations from reference baselines. The controlled comparison among Qwen3, Qwen3+LoRA, and SG-Layout isolates the effect of instruction tuning and graph conditioning under the same backbone family. The remaining methods are included as task-specific reference systems to position SG-Layout against prior LLM-, retrieval-, and generation-based pipelines.

\paragraph[Datasets.]{\textbullet\enspace Datasets.}
We use task-specific datasets for 2D, 3D, and rearrangement-style layout evaluation. NSR-1K \cite{feng2023layoutgpt} provides template-based and human-written prompts for testing spatial relation satisfaction. MSCOCO-Subset \cite{lin2014microsoft} is our curated 2D subset grouped by object count. 3D-FRONT/3D-FUTURE \cite{fu20213d} provides room layouts, furniture assets, and geometric annotations for indoor scene synthesis. SK-Dataset \cite{wang2025ske} is used to evaluate relational compliance under qualitative tabletop rearrangement instructions. The data scales are summarized in Table~\ref{tab:data_scale}.

\paragraph[Evaluation Metrics.]{\textbullet\enspace Evaluation Metrics.}
For image layout generation and object rearrangement,  we report AUC for spatial-relation
satisfaction, IoU for geometric overlap, and CLIP similarity for semantic consistency. For indoor scene synthesis, we evaluate structural accuracy, geometric accuracy, and physical feasibility. F1 measures category-level object matching after greedy one-to-one matching within each scene. Positional Error (Pos.$\downarrow$) and Rotational Error (Rot.$\downarrow$) are the mean center-distance and yaw-angle errors over matched objects. Out-of-Bound Rate (OOB$\downarrow$) is the fraction of predicted objects whose 3D boxes exceed the room boundary, and Collision Rate (Col.$\downarrow$) is the fraction of predicted objects colliding with at least one other object. Physically-Grounded Semantic Alignment (PSA$\uparrow$) measures the percentage of ground-truth directional relations preserved in the predicted layout according to the relation definitions in Eq.~(2).

\begin{table*}[h]
\vspace{-18pt}
\centering
\caption{Quantitative comparison on the indoor Scene Synthesis task.}
\begin{tabular}{lcccccc}
\toprule
\textbf{Models} & \textbf{PSA} & \textbf{Out of Bound} & \textbf{Collision} & \textbf{Pos.} & \textbf{Rot.} & \textbf{F1} \\
\midrule
DiffuScene & 25.3\% & -- & -- & -- & -- & -- \\
InstructScene & 58.3\% & -- & -- & -- & -- & -- \\
GPT-4 & 42.5\% & 76.3\% & 21.3\% & 40.86 & 80.20 & 0.973 \\
SKE-Layout (GPT-4) & 65.4\% & 48.7\% & \textbf{18.7\%} & \textbf{25.58} & 64.34 & \textbf{0.979} \\
SKE-Layout (Qwen3) & 35.2\% & 67.1\% & 21.2\% & 30.84 & 78.95 & 0.645 \\
\midrule
Qwen3 & 23.5\% & 80.9\% & 22.8\% & 37.67 & 82.47 & 0.563 \\
Qwen3 + LoRA & 56.7\% & 57.2\% & \underline{19.0\%} & 31.56 & 63.25 & 0.903 \\
SG-Layout(ours) & \textbf{66.2\%} & \textbf{43.5\%} & 19.2\% & \underline{26.89} & \textbf{57.45} & \underline{0.978} \\
\bottomrule
\end{tabular}

\vspace{2pt}
\footnotesize \emph{Bold} indicates the best overall result; \underline{underlined} values indicate the best Qwen3-based result when different.
\vspace{-50pt}
\label{tab:3droom_text_conditioned}
\end{table*}

\begin{table}[h]
\centering
\caption{Comparison of image layout accuracy under different numbers of objects}
\setlength{\tabcolsep}{3pt}
\renewcommand{\arraystretch}{1.08}
\footnotesize
\begin{tabularx}{\columnwidth}{l *{6}{>{\centering\arraybackslash}X}}
\toprule
\multirow{2}{*}{\textbf{Method}}
& \multicolumn{1}{c}{\textbf{K=2}}
& \multicolumn{1}{c}{\textbf{K=4}}
& \multicolumn{1}{c}{\textbf{K=6}}
& \multicolumn{3}{c}{\textbf{K=8}} \\
\cmidrule(lr){2-2}\cmidrule(lr){3-3}\cmidrule(lr){4-4}\cmidrule(lr){5-7}
& \textbf{AUC} & \textbf{AUC} & \textbf{AUC} & \textbf{AUC} & \textbf{IoU} & \textbf{CLIP} \\
\midrule
LayoutGPT (GPT-3.5) & 82.54 & ---  & ---  & ---  & --- & --- \\
LayoutGPT (GPT-4)   & 91.73 & 74.5 & 56.0 & 42.0 & 0.195 & 0.235 \\
SKE-Layout (GPT-4)  & \textbf{95.05} & \textbf{76.0} & \textbf{57.0} & \textbf{44.5} & \textbf{0.240} & \textbf{0.246} \\
SKE-Layout (Qwen3)  & 71.73 & 55.0 & 35.0 & 25.0 & 0.163 & 0.194 \\
\midrule
Qwen3               & 80.92 & 62.0 & 41.5 & 30.5 & 0.185 & 0.201 \\
Qwen3 + LoRA        & \underline{92.58} & \underline{73.0} & 49.0 & 38.0 & 0.205 & 0.218 \\
SG-Layout (ours)    & 89.85 & 71.5 & \underline{53.5} & \underline{42.5} & \underline{0.228} & \underline{0.240} \\
\bottomrule
\end{tabularx}

\vspace{2pt}
\footnotesize \emph{Bold} indicates the best overall result; \underline{underlined} values indicate the best Qwen3-based result.
\vspace{-25pt}
\label{tab:spatial_accuracy}
\end{table}

\subsection{Results and Discussions}
The overarching goal of our experiments is to test whether explicit graph-structured spatial knowledge improves a compact open-source LLM under controlled settings. Therefore, we emphasize comparisons within the Qwen3 family, where the backbone and task data are aligned, and use GPT-4-, retrieval-, and diffusion-based systems as reference baselines rather than strictly identical training protocols.
\paragraph[Image Layout Generation.]{\textbullet\enspace Image Layout Generation.}
Table~\ref{tab:spatial_accuracy} summarizes the quantitative results for the \emph{image layout generation} task, comparing how different models maintain spatial reasoning accuracy as scene complexity increases. As shown in the table, the performance of all models declines with more complex layouts, reflecting the growing difficulty of preserving spatial consistency under intricate object interactions.

Overall, GPT-4-based systems remain strong reference baselines, while the smaller Qwen3-8B backbone degrades more rapidly as the number of objects increases. Within the controlled Qwen3 family, LoRA tuning brings large gains on simpler layouts, whereas SG-Layout becomes more beneficial as relation density increases. In particular, SG-Layout gives the best Qwen3-based results at $K{=}6$ and $K{=}8$, as well as higher IoU and CLIP scores than the text-only LoRA variant, suggesting that explicit graph conditioning is most useful for compositional spatial layouts.

\paragraph[Indoor Scene Synthesis.]{\textbullet\enspace Indoor Scene Synthesis.}
Table~\ref{tab:3droom_text_conditioned} summarizes the quantitative results for the \emph{3D indoor scene synthesis} task, evaluating each model’s semantic alignment and physical plausibility in generating realistic and coherent room layouts. As shown in the table, SG-Layout achieves substantial improvements over the Qwen3 backbone and its variants, demonstrating more consistent and physically grounded 3D spatial reasoning.

SG-Layout achieves the highest PSA and the lowest OOB rate in Table~\ref{tab:3droom_text_conditioned}, showing that graph embeddings help preserve spatial relations and room-boundary constraints. Within the Qwen3-based controlled setting, SG-Layout also improves Pos., Rot., and F1 over the vanilla and text-only LoRA variants, although Qwen3+LoRA has a slightly lower collision rate. These results support our central claim that graph conditioning improves a fixed compact backbone, while the comparison with GPT-4/SKE-Layout should be interpreted as a reference comparison rather than a fully identical training protocol.

\begin{figure}[t]
  \centering
  \small

  \begin{minipage}[c]{0.48\linewidth}
    \centering
    \vspace{0pt}
    \setlength{\tabcolsep}{1pt}
    \renewcommand{\arraystretch}{1.05}
    \begin{tabular}{lcccc}
      \toprule
      \textbf{Encoder} & \textbf{PSA$\uparrow$} & \textbf{OOB$\downarrow$} & \textbf{Collision$\downarrow$} & \textbf{F1$\uparrow$} \\
      \midrule
      R-GCN & 63.8\% & 45.9\% & \textbf{18.1\%} & 0.974 \\
      RGT   & \textbf{66.2\%} & \textbf{43.5\%} & 19.2\% & \textbf{0.978} \\
      GAT   & 64.3\% & 47.7\% & 18.5\% & 0.965 \\
      \bottomrule
    \end{tabular}
    \captionof{table}{Graph encoder ablation on 3D indoor scene synthesis.}
    \label{tab:encoder_ablation_3d}
  \end{minipage}\hfill
  \begin{minipage}[c]{0.48\linewidth}
    \centering
    \vspace{0pt}
    \includegraphics[width=\linewidth]{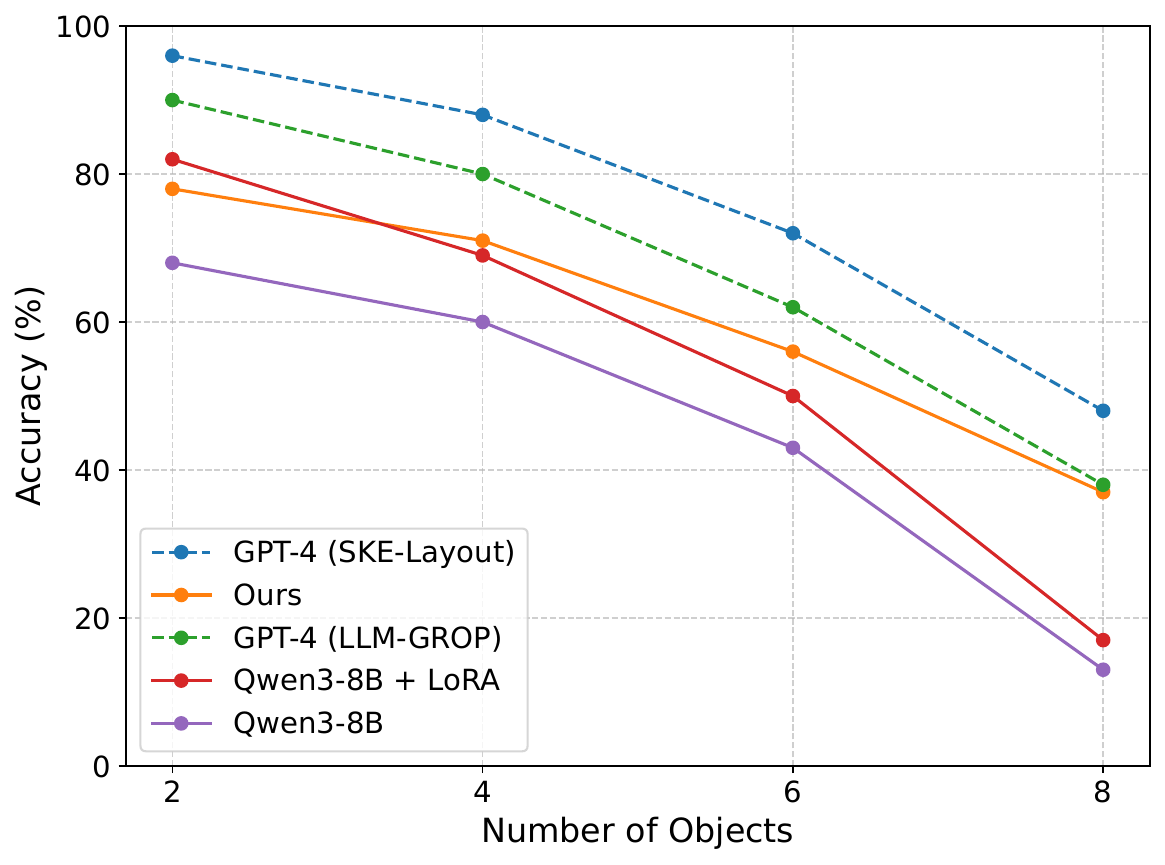}

    \captionof{figure}{3D Object Rearrangement Layout Accuracy.}

    \label{fig_3}
  \end{minipage}
  \vspace{-20pt}
\end{figure}

\paragraph[Object Rearrangement.]{\textbullet\enspace Object Rearrangement.}
Figure~\ref{fig_3} reports layout accuracy for the object rearrangement task under different numbers of objects. The LoRA-enhanced Qwen3-8B performs well in simple configurations but degrades as the scene becomes more crowded. SG-Layout maintains higher accuracy in higher-object settings, indicating that graph-structured conditioning helps preserve qualitative spatial relations when instructions involve multiple interacting objects. GPT-4-based systems such as LLM-GROP and SKE-Layout remain strong reference baselines, but SG-Layout provides a parameter-efficient alternative built on a compact open-source backbone.

\paragraph{Ablation on Graph Encoders.}
To study the impact of graph encoding architectures, we instantiate the scene-graph encoder with three alternatives: (i) R-GCN \cite{schlichtkrull2018modeling} as a lightweight relational baseline, (ii) GAT \cite{velivckovic2017graph} with relation-type embeddings, and (iii) a relational graph transformer (RGT)\cite{dwivedi2025relational} that models typed relations via attention.
As shown in Table~\ref{tab:encoder_ablation_3d}, transformer-style relational encoding (RGT) yields the most consistent gains, especially on constraint-sensitive metrics.
These results suggest that improving relation modeling in the graph encoder directly benefits spatial grounding and compositional consistency in SG-Layout.

\begin{figure}[h]
\captionsetup{aboveskip=2pt}
\centering
\includegraphics[width=0.98\linewidth]{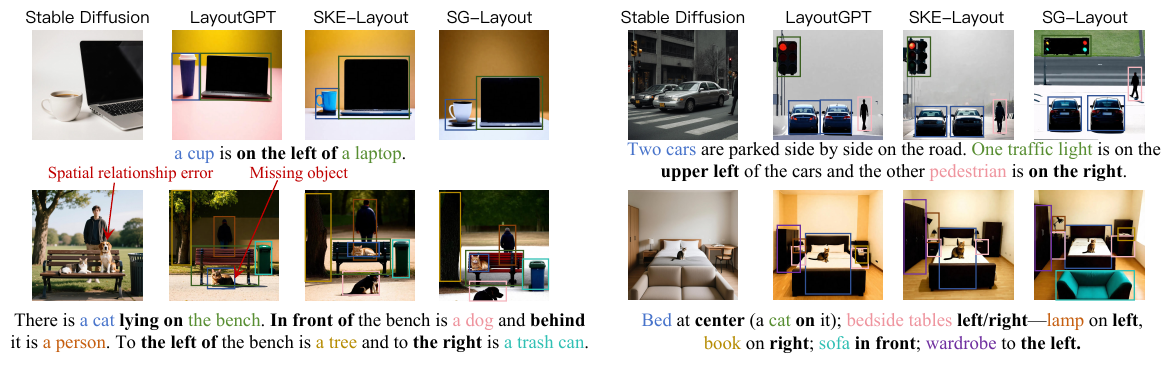}
\caption{Comparison results across prompts with increasing object counts on image layout generation.}
\label{2dcase}
\captionsetup{belowskip=4pt}
\end{figure}

\begin{figure}[h]
\captionsetup{aboveskip=2pt}
\centering
\includegraphics[width=0.98\linewidth]{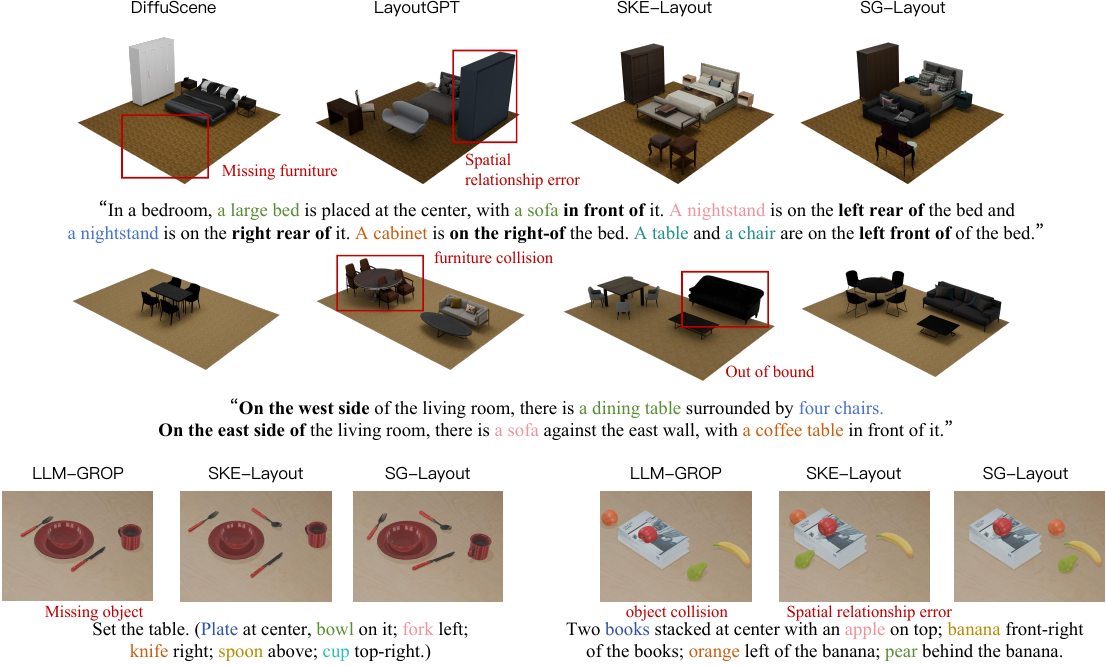}
\caption{Comparison results across scenes with varying relations on indoor scene synthesis and 3D object rearrangement tasks.}
\vspace{-20pt}
\label{3dcase}
\end{figure}
\subsection{Case Study}

To complement quantitative metrics and make failure modes visually explicit, we present several case studies. Comparisons in Fig.~\ref{2dcase} (2D) and Fig.~\ref{3dcase} (3D) illustrate typical failure modes of prior methods and how SG-Layout overcomes them. In 2D  \emph{image layout generation}, most methods succeed on simple two-object prompts, but as object count and relational density increase they exhibit missing or duplicated objects and violated relations, as shown in Fig.~\ref{2dcase}. In 3D \emph{indoor scene synthesis} and \emph{object rearrangement}, baselines frequently show missing furniture, inter-object collisions, out-of-bound placements, and incorrect spatial relationship, as shown in Fig.~\ref{3dcase}. Taken together, these failure modes are mitigated by \textbf{SG-Layout}, which conditions decoding on scene-graph–derived structured guidance via a two-stage alignment and instruction-tuning scheme, yielding more accurate and consistent layouts with fewer errors in both 2D and 3D layouts.

\section{Conclusion}
In this work, we introduce \textbf{SG-Layout}, a graph-guided LLM designed for text-conditioned 2D and 3D layout generation.
By decomposing the training process into two stages, graph–language feature alignment and LoRA-based instruction tuning, SG-Layout effectively bridges structured spatial representations with the linguistic latent space of the LLM.
Extensive experiments on both image and indoor scene benchmarks show that this approach enhances spatial reasoning accuracy, reduces physical violations, and yields layouts that are both semantically and geometrically consistent. Compared to the original LLM backbone, SG-Layout achieves clear improvements in spatial coherence while maintaining parameter efficiency. Looking ahead, we will target precisely quantifiable domains—e.g., surgical-tool layout on trays and in operating rooms—where millimeter-level tolerance constraints must be satisfied.
We will also explore reinforcement feedback and human-in-the-loop refinement for LLMs.

\begin{credits}
\subsubsection{\ackname}
This work was supported by the National Natural Science Foundation of China under Grants 62322601 and 62572084, and the Fundamental Research Funds for the Central Universities (Nos. 2024IAIS-QN017 and 2025CDJZDGF001).

\subsubsection{\discintname}
The authors have no competing interests to declare that are relevant to the content of this article.
\end{credits}

\bibliographystyle{splncs04}
\bibliography{main}
\end{document}